# 3D-TexSeg: Unsupervised Segmentation of 3D Texture using Mutual Transformer Learning

Iyyakutti Iyappan Ganapathi
Khalifa University
Abu Dhabi, UAE.
iyyakutti.ganapathi@ku.ac.ae

Fayaz Ali
Khalifa University
Abu Dhabi, UAE.
fayaz.ali@ku.ac.ae

Sajid Javed
Khalifa University
Abu Dhabi, UAE.
Sajid.Javed@ku.ac.ae

Syed Sadaf Ali
Khalifa University
Abu Dhabi, UAE.
syed.ali@ku.ac.ae

Naoufel Werghi
Khalifa University
Abu Dhabi, UAE.
naoufel.werghi@ku.ac.ae

## Abstract

*Analysis of the 3D Texture is indispensable for various tasks, such as retrieval, segmentation, classification, and inspection of sculptures, knitted fabrics, and biological tissues. A 3D texture is a locally repeated surface variation independent of the surface's overall shape and can be determined using the local neighborhood and its characteristics. Existing techniques typically employ computer vision techniques that analyze a 3D mesh globally, derive features, and then utilize the obtained features for retrieval or classification. Several traditional and learning-based methods exist in the literature; however, only a few are on 3D texture, and nothing yet, to the best of our knowledge, on the unsupervised schemes. This paper presents an original framework for the unsupervised segmentation of the 3D texture on the mesh manifold. We approach this problem as binary surface segmentation, partitioning the mesh surface into textured and non-textured regions without prior annotation. We devise a mutual transformer-based system comprising a label generator and a cleaner. The two models take geometric image representations of the surface mesh facets and label them as texture or non-texture across an iterative mutual learning scheme. Extensive experiments on three publicly available datasets with diverse texture patterns demonstrate that the proposed framework outperforms standard and SOTA unsupervised techniques and competes reasonably with supervised methods.*

## 1. Introduction

With the widespread 3D cameras and scanning devices capturing rich geometrical properties of object surfaces, many computers vision-based interdisciplinary applications have emerged in recent years. A large volume of work has been addressing the problem of segmenting, classifying, and retrieving 3D shapes based on their similarities using triangle mesh and point clouds as input [5, 13, 15, 34]. However, a less investigated but emerging problem is the segmentation and classification of 3D geometric texture (or simply 3D texture). The 3D texture is a surface feature different from the shape and is characterized by repetitive geometric regular or random patterns on the surface. These patterns can be viewed as geometric corrugations of the surface altering the local smoothness and appearance of the surface, however, without affecting its global shape. A large variety of surfaces exhibiting 3D texture, which include knitted fabrics, artwork patterns, artist styles, and natural structures like tree barks [23, 37]. Several industries, including remote sensing, 3D content creation, and animation, can benefit tremendously from texture-based applications [1]. Cultural preservation is one among these, where cultural object retrieval and categorization based on texture have been the subject of extensive research and development [8, 10, 11, 22]. Recent advances in the field have shown remarkable performance in transforming historical buildings into semantically structured 3D models, enabling enhanced detection and comprehension of heritage structures [20].

All the 3D texture classification and segmentation methods developed so far have relied on supervised schemes that require demanding manual annotation of a large amount of data. Manual annotation of textured regions on 3D surfaces is even more tedious than its counterpart in 2D images as it requires repeating the procedure over multiple views. Also, the manual annotation is susceptible to systematic error because the annotator operates on a 2D projection of the sur-face.

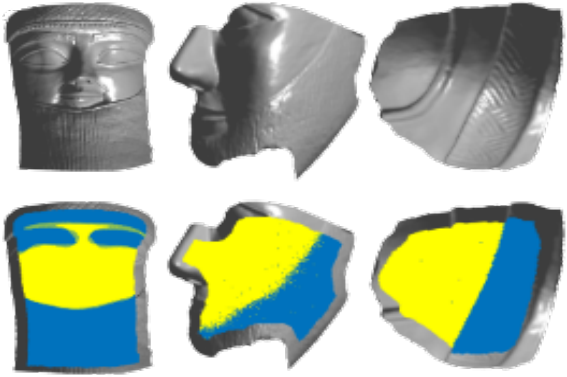

Figure 1. 3D surface samples with varied texture patterns. The top row depicts cultural heritage artifacts, while the bottom row depicts segmented regions, with yellow indicating non-texture and blue indicating texture.

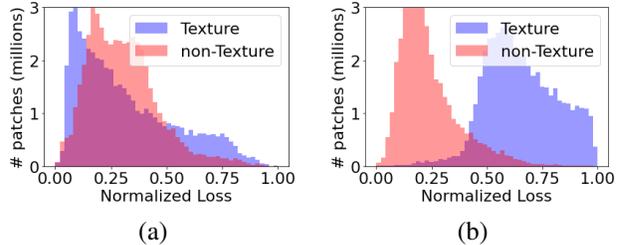

Figure 2. Distributions of reconstruction loss for texture and non-texture patches. (a) Losses obtained early in the process show that there is a significant overlap between the distributions of texture and non-texture patches, resulting in a high misclassification error, whereas (b) losses obtained near the end of the process show that there is a noticeable separation between the distributions of texture and non-texture patches, resulting in a lower misclassification error.

In this paper, we present an original framework for the unsupervised segmentation of the 3D texture segmentation on the mesh manifold. The problem is approached as a fully unsupervised binary surface segmentation where the mesh surface is partitioned into textured and non-textured regions (see examples in Figure 1). This novel scheme eliminates labor-intensive labeling while achieving comparable segmentation performance to the supervised methods. To our knowledge, this is the first attempt to address such a problem.

Our approach is inspired by observing the behavior of autoencoder models when we used them to reconstruct surface patches. We discovered that the reconstruction error for a textured patch (heterogenous) is often greater than its counterpart in the non-textured patch (homogeneous or smooth patched). In Figure 2, we report the distribution of the reconstruction error for two sets of textured and non-textured patches collected from different surfaces. This disparity can be explained by the heterogeneity of the textured surface, thus presenting a larger entropy compared to the homogenous non-textured patches. From these observations, we hypothesize that this behavior's disparity could be accentuated further and leveraged via a cleaner learning mechanism within an adversarial scheme for fully unsupervised classification of the surface patches.

The proposed model includes a label generator and a cleaner. The generator is trained to reconstruct surface patch features. The reconstruction loss function is used to label the patch, whereby large loss and low loss patches are assigned to non-texture and texture classes, respectively. This set of pseudo-labels is excepted to contain several misclassified patches, and thus there is a need for further segregation. For this purpose, we introduce a discriminative learning mechanism in which a binary classifier is trained with the pseudo-labeled patches and then used to reclassify the patches in the second stage, correcting the initial assignment. For example, a patch initially labeled as textured can be reclassified as non-textured and vice-versa. This scenario is quite possible since classifier training is never expected to be 100% accurate. The modified set of pseudo-labeled patches is then utilized in the second iteration to enhance the generator further. By iterating this procedure, the pseudo-label generator and pseudo-label cleaner modules mutually learn from each other and improve the overall surface patch classification performance.

The proposed framework outperforms the classical unsupervised approaches and baseline methods on three datasets: *KU 3DTexture* [7], *SHREC'18* [3], and *SHREC'17* [2]. In summary, our original contributions are summarized as follows:

1. We propose leveraging the surface patch reconstruction error as an underlying concept for classifying textured and non-texture patches.
2. We present a fully unsupervised mutual transformer learning approach for 3D texture segmentation on mesh surfaces. To the best of our knowledge, this is the first attempt at facet-level texture segmentation.
3. We validate the proposed framework for texture segmentation on three datasets with complex texture patterns and varying resolutions, achieving significantly better results than conventional clustering and baseline approaches.

## 2. Related Work

As a recent topic, there is not yet a large volume of work on 3D texture analysis. Nonetheless, the research developed so far can be categorized into 3D texture classification [31–33], 3D texture retrieval [2, 27, 28], and 3D texture segmentation [17, 29, 36]. In the classification, Werghi *et al.*

[31] pioneered the concept of 3D texture. They proposed the mesh-LBP as an extension of the local binary pattern to the mesh manifold, using a structure of local ordered rings, and applied it to classify textured patterns on mesh surfaces. In subsequent work [32, 33], they extended it to other applications such as 3D face recognition. Later, this newly proposed 3D texture concept attracted the attention of the SHape REtrieval Community, which released a series of 3D relief pattern datasets in the SHREC contests [2]. Moscoso et al. proposed the edge-LBP descriptor using contours defined based on a sphere-mesh intersection and employed this representation for matching archaeological fragments using Battacharya distance as metric [27]. In [28] Thompson et al. reported a variety of techniques, all to revolve around the best representation characterizing 3D texture patterns and related similarity metrics.

Segmentation of 3D textures on the mesh is still in its infancy stage. In [17], Lie et al. proposed a supervised snake-based segmentation approach. The method requires a manual selection of snake contours which then evolve towards separating the smooth surfaces and the relief patterns. Zatzarinni et al. [36] addressed similar problems analytically using a height function defined over the surface. These methods are meant to treat relief patterns, characterized by their protrusion over the main surface, and cannot generalize to the 3D texture. More recently, Tortoricci et al. [29] proposed convolution tools to extract texture features on the mesh, which are employed in a weakly supervised scheme using Random Forest. In [4], Choi et al. proposed a semantic segmentation approach using FC-DenseNet to extract 3D scripts from rough surfaces. The model is trained with feature images constructed from local shape features.

## 3. Proposed Methodology

The schematic illustration of our proposed method is depicted in Figure 3. The method encompasses three main steps: patch image extraction, deep features extraction, and unsupervised patch classification.

### 3.1. Surface patch image extraction

Our segmentation technique uses local classification, in which the mesh surface is browsed, and a neighborhood around each triangle facet is constructed; from each neighborhood creates a multichannel geometric image with each channel representing a geometric feature. The multichannel image is constructed using the ordered ring facets (ORF) structure developed in [32]. We extract an ORF from each facet and utilize it to generate a grid to encode facets as a 2D matrix. Further, at each facet, three geometric descriptors are computed: local depth, surface variation, and mean curvature, and the resulting geometric maps are stacked to generate a 3-channel geometric image which we refer to as the *surface patch image* as shown in Figure 4.

### 3.2. Deep feature extraction

The geometric image, while reflecting the local geometry of a surface patch, does not possess sufficient discrimination capacity. For improved discrimination, a pre-trained restNet model is employed to create a deep feature representation $f$, from geometric images. The model has not been tuned or exposed to texture or non-texture data in an effort to stick to the concept of a completely unsupervised framework.

### 3.3. Unsupervised patch classification

As mentioned before, our unsupervised patch classification employs a model composed of two modules, the label generator, and the label cleaner. The two models encompass an autoencoder-like model and a binary classifier, respectively. For both models, we adopted a transformer backbone architecture. While transformers demonstrated remarkable performance in several image analysis tasks [12, 14, 38], our primary motivation stems from their capacity to model both short-range and long-range dependencies. This aspect is quite present in the textured surface patches because of the repetitive patterns all along their surface. We dubbed the label generator and the label cleaner the Transformer Label Generator (TLG) and the Transformer Label Cleaner (TLC).

### 3.4. Initial Patch Clustering (IPC)

Unsupervised learning techniques are more effective when class samples are homogeneous and compact (e.g., a k-mean clustering works fine when the feature space's class distributions are compact and reasonably separated). While such an ideal scenario is unlikely in our data, we can reduce the heterogeneity of the classes' samples (here, patch instances in the texture and the non-texture classes). Assuming our deep features have adequate discrimination capacity, one method is to do mean-shift clustering on the deep feature samples, select the two most predominant clusters, and discard the rest. The two dominating clusters are anticipated to display reasonable compactness as a density-based approach, whereas the excluded clusters are most likely to contain hard samples. Another simpler and computationally less demanding approach, which we found working reasonably, is to run the *K-means* clustering with many clusters above 2. In the experimentation, we empirically found $K = 5$, a suitable value.

**Transformer Label Generator (TLG)** Our transformer projector comprises a Multi-head Self Attention (MSA) layer and a Multi-Layer Perceptron (MLP) containing two fully connected layers. The filtered patch instances obtained from the IPC are passed to TLG. Here, their deep feature representations are projected into a latent space using a transformer-based projector and then inverse-transformed to the original space using a transformer-based inverse projector. The transformation loss is then used to assign

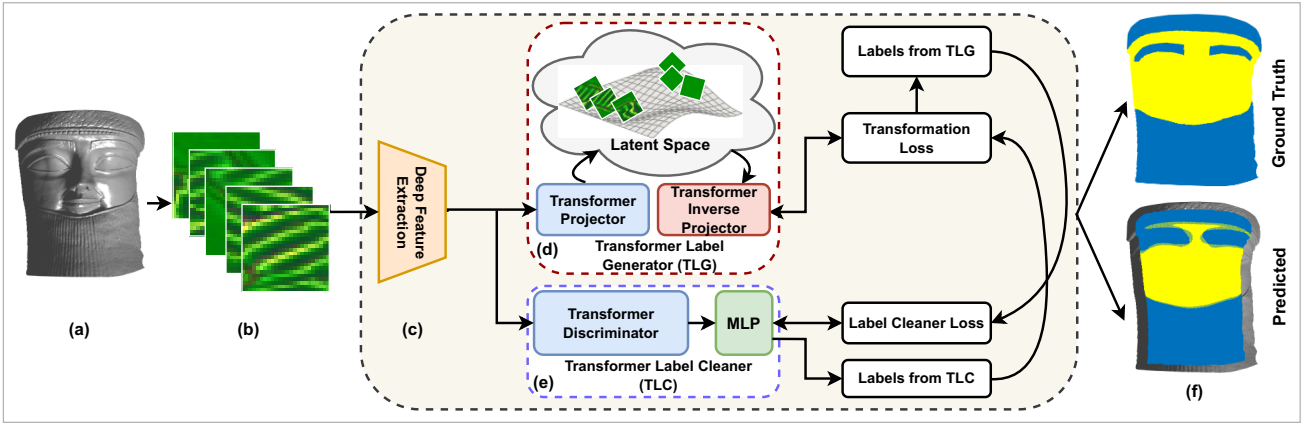

Figure 3. Outline of the proposed surface patch classification for texture segmentation. (a) 3D surface, (b) 2D surface patch images computed across the mesh triangle facets using geometric features (see Figure 4). (c) Deep feature extraction from the surface patch images. (d) a label generator inputs deep features and assigns a pseudo-label (texture or non-texture) to each surface patch. Noticeably, this assignment produces misclassified surface patches (i.e., noisy labels). (e) a label cleaner cleans the noisy pseudo-labels generated in (d) repeated over several iterations, and (f) ground truth and the predicted results, where yellow and blue represent the non-texture and texture regions, respectively.

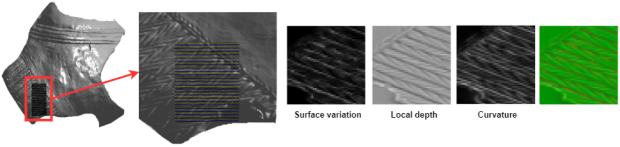

Figure 4. An example of a surface patch image extraction. A facet grid is constructed around a central facet (here, a 24 × 24), and three different geometric descriptors are computed at each facet of the grid: surface variation, local depth, and curvature, producing a three-channel image.

pseudo-labels to each patch instance.

We employed a similar transformer architecture proposed by Vaswani *et al.* [30]. Let $N$ be the number of patches in the mesh surface, and let $f_i$ be the deep feature representation of the $i^{th}$ patch, then we re-arrange $f_i$ as a sequence of position-aware word representations $g_i = [g_{i,1}, g_{i,2}, ..., g_{i,n_k}]$, $n_k$ is the length of the sequence. The projector converts $g_i$ to a latent representation $p_L$ via the following sequence of transformations:

$$\begin{aligned} p_0 &= g_i, \\ q_x = k_x = v_x &= \mathbf{LN}(p_{x-1}), \\ \hat{p}_x &= \mathbf{MSA}(q_x, k_x, v_x) + p_{x-1}, \\ p_L &= [q_{i,1}, q_{i,2}, ..., q_{i,n_k}] \end{aligned} \quad (1)$$

where $x = 1, ..., L$ denotes the number of layers and $LN$ represents Layer Normalization. In the TLG architecture, the latent space retains the same size as the input sequence.

The architecture of the inverse projector is similar to that of the transformer projector. It consists of two MSA layers followed by MLP. There is also a latent learned bias vector $b$ utilized in reconstructing features $z_L = [\hat{g}_{i,1}, \hat{g}_{i,2}, ..., \hat{g}_{i,n_k}]$ via the sequence of transformations

$$\begin{aligned} z_0 &= p_L, \\ q_x = k_x &= \mathbf{LN}(z_{x-1}) + b, v_x = \mathbf{LN}(z_{x-1}), \\ \hat{z}_x &= \mathbf{MSA}(q_x, k_x, v_x) + z_{x-1}, \hat{q}_x = \mathbf{LN}(\hat{z}_x) + b, \\ \hat{k}_x = \hat{v}_x &= \mathbf{LN}(z_0), \tilde{z}_x = \mathbf{MSA}(\hat{q}_x, \hat{k}_x, \hat{v}_x) + \hat{z}_x, \\ z_x &= \mathbf{MLP}(\mathbf{LN}(\tilde{z}_x)) + \tilde{z}_x \end{aligned}$$

We optimize the TLG by minimizing the following loss function:

$$\mathcal{L}_{TLG} = \sum_{i=1}^{n} ||g_i - \hat{g}_i||_1 \quad (2)$$

where $n$ is the total number of surface patches in a batch. Once optimized, the reconstruction error is computed for the $i^{th}$ patch instance as

$$e^i_{TLG} = ||g_i - \hat{g}_i||_1 \quad (3)$$

Afterward, we generate its pseudo-label in the first iteration by thresholding,

$$l_i = \begin{cases} 1 \; if & e^i_{TLG} - average_{batch}(e^i_{TLG}) \geq 0 \\ 0 & \text{otherwise} \end{cases} \quad (4)$$

where the label 1 and 0 correspond to the texture and non-texture, respectively.

In the subsequence iterations, the pseudo-label assignment is modified. For a patch labeled non-texture in the

**Algorithm 1** Proposed algorithm
---
**Require:** N deep feature vectors representing all the surface patches
  **for** each epoch **do**
    Take a batch of *n* samples
    Minimize TLG's loss function (2)
    Compute construction errors $e^i_{TLG}$, i=*1:n*, using equation (3)
    set pseudo-label $l_i$ using equation (4)
    ———Label Cleaner———
    Minimize TLC's loss function (6) using the labels $l_i$
    Compute $\phi_i \leftarrow TLC(g_i, l_i), \; i = 1 : n$
    **if** $\phi_i > \beta_c$ **then**
      $l_i \leftarrow$ textr
    **end if**
    **while** iter **do**
      Take a batch of *n* samples
      Minimize TLG's loss function (2)
      **if** $l_i$ == textr **then**
        $g_i \leftarrow Gaussian\;noise(g_f)$
        Compute $e^i_{TLG}$ as per equation (5)
      **else**
        Compute $e^i_{TLG}$ as per equation (3)
      **end if**
      ———Label Cleaner———
      Minimize TLC's loss function (6) using the labels $l_i$
      Compute $\phi_i \leftarrow TLC(g_i, l_i), \; i = 1 : n$
      Compute $l_i$ using equation (7)
    **end while**
  **end for**
  **Return** cleaned label $l_i$

---

previous iteration, the reconstruction error of equation (3) is used. The reconstruction error of the following equation is used for a patch-labeled texture.

$$e^i_{TLG} = ||\tilde{g}_f - \hat{g}_i||_1 \quad (5)$$

where $\tilde{g}_f$ is a random Gaussian vector having normal distribution. Empirically, we found that switching to the above formula enhances the capacity of the TLG to detect the textured patches and improves the overall segmentation. To train a generator to produce desired images in a generative framework, a negative correlation between the discriminator and generator losses must be achieved [9]. In our network, a similar approach is employed to get desired labels by increasing the loss of the discriminator for texture by providing a fixed gaussian as input.

**Transformer Label Cleaner (TLC):** We also employ transformer architecture similar to the transformer projector for the TLC, where the last layer is connected to a dense neuron. Further, the TLC, a binary classifier, is trained with the patches used in the previous step and their pseudo-labels generated in (4), using a simple binary cross-entropy loss

$$\mathcal{L}_{TLC} = \frac{1}{n}\sum_{i=1}^{n} -(l_i log(\phi_i) + (1 - l_i) log(1 - \phi_i)) \quad (6)$$

where $\phi_i$ is the output of the binary classifier represents the probability of a textured region, and $1 - \phi_i$ represents the probability of a non-textured region. Once trained, each patch instance is passed to the binary classifier, and its label is adjusted as follows:

$$l_i = \begin{cases} 1 \; \text{if} \phi_i \geq average_{batch}(\phi_i), \\ 0 \; \text{otherwise}, \end{cases} \quad (7)$$

These adjusted labels $l_i$ are used to train the label generator in the next iteration.

TLG and TLC alternate over the batch of surface patches until the mesh surface is completely covered. The algorithm goes into the next epoch till a maximum number of epochs is reached. Figure 5 depicts an exemplar of the evolution of the patch classification across the epochs. It is evident that the segmentation improves as the number of iterations increases, resulting in well-separated textured and non-textured regions.

## 4. Experimental Results

We evalaute our frameworks using three datasets: *SHREC'17* [2], *SHREC'18* [3], and *KU 3DTexture* [7]. *SHREC'17* contains 15 distinct textures with 720 meshes, and each texture class contains 48 samples with varying mesh resolutions. The dataset *SHREC'18* has twelve distinct surfaces with distinct texture patterns, each with a unique resolution. The *Ku 3DTexture* [7] contains 89 real-world data samples with dense texture regions. Since the problem involves classifying each facet, the number of facets exposed to the network is essential. The data has a minimum of 10K to 785K facets per sample. Even though the number of surfaces used is small, we found that the overall number of available facets is sufficient to train the network. Despite this, we have utilized augmented data and subjected our model to various surface variances to generalize to previously unseen patches.

The performance of the proposed method is compared to nine existing techniques, including six techniques based on deep learning and three conventional unsupervised techniques. The proposed method is evaluated and compared using F1-Score, Precision, and Recall, with an IoU threshold of 0.5. Additionally provided is the mean Intersection over Union (mIoU) score. The objective is to categorize each facet of a given surface as belonging to a texture or non-texture region.

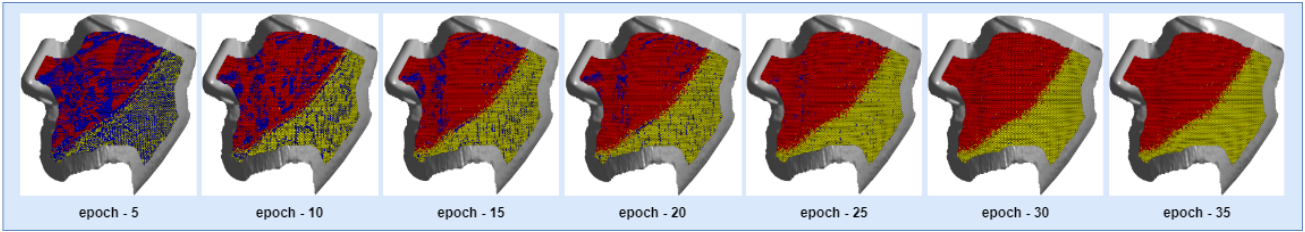

Figure 5. An illustration of the segmentation improvement across the iterations. Correctly classified texture facets are colored in yellow, non-texture in red, and misclassified in blue.

## 4.1. Quantitative Analysis

We compared the proposed method to unsupervised and fully supervised approaches. Since there is no current unsupervised approach for texture segmentation on 3D surfaces, we initially implemented three traditional methods to compare the proposed method with: *K-Means* [18], *DBSCAN* [6], and *GMM Clustering* [21]. In addition, we compare the performance of the proposed method with popular 3D shape classification segmentation networks [16, 19, 24–26, 35]. The objective of these networks is to segment distinctive and consistent structures. In particular, the shape is employed to distinguish the unique structures of each class. In our case, the objective is to classify each point or facet as textured or non-textured based on surface variations in a small region instead of segmenting the shape. Additionally, there is a disparity between the proportion of texture and non-texture regions. As a result, we updated the loss functions and incorporated balanced focal loss, which forces the model to concentrate on challenging classes while adapting these models to texture/non-texture classification. As input, we utilized point clouds or 3D meshes with each point and facet labeled.

### 4.1.1 Evaluation on *KU 3DTexture* dataset

The results of our method, together with other sets of supervised and unsupervised approaches, are reported in Table 1. Our method surpasses the classical clustering-based unsupervised methods [6, 18, 21] by large margins on all metrics showing the advantage of our method in fully leveraging both transformer modules, label generator, and cleaner. The mutually trained transformer modules have improved the segmentation accuracy compared to classical techniques, and the results obtained using the proposed unsupervised is close to the supervised method.

The proposed unsupervised approach has superior performance than four techniques [16, 24–26] out of six. *KU 3DTexture* has diverse patterns and complex surfaces, so the results obtained are slightly less compared to the other two datasets. Also, in the case of the supervised approach, the proposed approach has shown superior performance com-pared to [16, 19, 24–26, 35]. Though these approaches are designed for 3D shape analysis and demonstrated remarkable performance on semantic segmentation of 3D shapes, in our case, they failed in capturing the textures on 3D surfaces.

Table 1. Quantitative results of our method and baselines on the *KU 3DTexture* [7] dataset

| | | Pre ↑ | Rec ↑ | F1 ↑ | mIoU ↑ |
|---|---|---|---|---|---|
| Supervised Approaches | PointNet [CVPR'17][24] | - | - | - | 48.2 |
| | PointNet++ [NeurIPS'17] [25] | - | - | - | 49.7 |
| | MeshSegNet [TMI'20] [16] | - | - | - | 58.0 |
| | BAAFNet [CVPR'21] [26] | - | - | - | 51.1 |
| | PointMLP [ICLR'22] [19] | - | - | - | 67.0 |
| | CurveNet [ICCV'21] [35] | - | - | - | 64.0 |
| | Proposed$_{sup}$ | **75.7** | **80.6** | **78.1** | **80.3** |
| Unsupervised Approaches | K-Means [18] | 12.6 | 18.4 | 16.1 | 22.5 |
| | DBSCAN [6] | 17.1 | 26.8 | 22.5 | 30.6 |
| | GMM Clustering [21] | 6.9 | 10.1 | 23.5 | 15.3 |
| | **Proposed** | **65.2** | **66.4** | **65.0** | **66.2** |

### 4.1.2 Evaluation on *SHREC'17* dataset

This dataset presents a significant challenge due to the wide variety of mesh resolutions and texture patterns. The proposed technique has yielded promising results and demonstrates its robustness against varying mesh resolutions. Table 2 clearly shows that the proposed method under supervised and unsupervised conditions performed better than the classical and deep learning-based methods. Moreover, it is worth mentioning that the proposed unsupervised approach performs better than all the supervised approaches [16, 19, 24–26, 35] with a better margin. The proposed method using supervised and unsupervised is the best performer, and pointMLP [19] and DBSCAN [6] are the second best performer.

### 4.1.3 Evaluation on *SHREC'18* dataset

We additionally evaluate our method on SHREC'18, which has 3D surfaces with multiple texture patterns on each surface with complex boundaries between the patterns. Also, the surfaces with varying mesh resolution which is further challenging. As shown in Table 3, the proposed method under supervised and unsupervised is the best performer, and curveNet [35], and K-Means [18] is the second best performer. Also, the scores obtained by our unsupervised approach are close to our fully supervised counterpart, and also it is superior to all supervised approaches [16, 19, 24–26, 35] except [35].

Table 2. Quantitative results of our method and baselines on the *SHREC'17* [2] dataset

| | | Pre ↑ | Rec ↑ | F1 ↑ | mIoU ↑ |
|---|---|---|---|---|---|
| Supervised Approaches | PointNet [CVPR'17] [24] | - | - | - | 51.8 |
| | PointNet++ [NeurIPS'17] [25] | - | - | - | 48.3 |
| | MeshSegNet [TMI'20] [16] | - | - | - | 62.4 |
| | BAAFNet [CVPR'21] [26] | - | - | - | 56.2 |
| | PointMLP [ICLR'22] [19] | - | - | - | 67.3 |
| | CurveNet [ICCV'21] [35] | - | - | - | 66.4 |
| | Proposed$_{sup}$ | **81.4** | **80.0** | **82.1** | **79.0** |
| Unsupervised Approaches | K-Means [18] | 23.6 | 21.4 | 22.6 | 30.5 |
| | DBSCAN [6] | 27.1 | 26.4 | 26.7 | 36.5 |
| | GMM Clustering [21] | 12.1 | 8.3 | 10.2 | 16.4 |
| | **Proposed** | **68.2** | **69.1** | **69.0** | **70.1** |

Table 3. Quantitative results of our method and baselines on the *SHREC'18* dataset [3]

| | | Pre ↑ | Rec ↑ | F1 ↑ | mIoU ↑ |
|---|---|---|---|---|---|
| Supervised Approaches | PointNet [CVPR'17] [24] | - | - | - | 54.2 |
| | PointNet++ [NeurIPS'17] [25] | - | - | - | 58.1 |
| | MeshSegNet [TMI'20] [16] | - | - | - | 60.3 |
| | BAAFNet [CVPR'21] [26] | - | - | - | 58.7 |
| | PointMLP [ICLR'22] [19] | - | - | - | 66.7 |
| | CurveNet [ICCV'21] [35] | - | - | - | 70.4 |
| | Proposed$_{sup}$ | **86.2** | **85.4** | **84.3** | **82.0** |
| Unsupervised Approaches | K-Means [18] | 33.6 | 25.4 | 29.5 | 38.2 |
| | DBSCAN [6] | 28.7 | 30.6 | 29.4 | 35.0 |
| | GMM Clustering [21] | 10.3 | 8.2 | 9.1 | 12.1 |
| | **Proposed** | **68.1** | **69.6** | **70.0** | **73.4** |

### 4.2. Qualitative Analysis

A few samples in Figure 6 show the effectiveness of the proposed technique. The 3D surfaces presented have various texture patterns; many texture patterns are visible even within one surface. Since we are interested in binary classification, we consider all texture patterns one class and all non-texture patterns another. A few facet misclassifications on a few 3D segmented surfaces using qualitative analysis are discovered, particularly at the texture and non-texture boundaries. This is because using ordered rings around a facet at boundaries covers neighboring facets from texture and non-texture regions. We use a wide range of facets, from texture and non-texture, to handle these challenges to some extent. However, the issues are inescapable because the surfaces come in various patterns and resolutions. Middle and the bottom rows in Figure 6 shows the ground truth of surfaces and the predicted results, where blue represents the texture region and yellow represents the non-textured region. Since ORF does not cover the entire surface because of boundary restrictions, only the central portion of surfaces are utilized for training and testing, as shown in the third row of Figure 6.

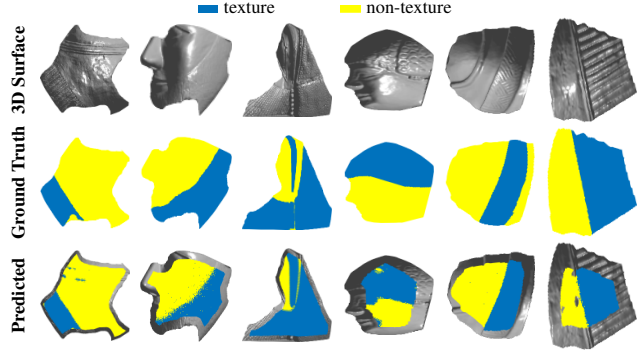

Figure 6. Visualization of segmentation of a few samples using the proposed technique. The top row is the original 3D surfaces, the middle row is the ground truth, yellow represents the non-texture region, blue represents the texture region, and the third row is the predicted facet level segmentation.

## 5. Ablation Studies

We conducted five ablative tests on the *SHREC'18* dataset to assess variants of the proposed model. We examine the performance of a network by removing certain components to determine each component's contribution to the overall system, as shown in Table 5. The performance of variants reflects the relative contributions of each component. 1) Remove only the *IPC*, the overall F1-score is degraded by 4%, 2) Remove only the *label cleaner*, the overall F1-score reduce by 6%, 3) Remove only the *label generator*, the overall F1-score is reduced by 8.5%, and 4) Remove only DL, the overall F1-score is reduced by 1.5%. Combining the proposed modules yields the highest segmentation scores, as shown in Table 5. Therefore, even though *IPC* does not precisely label the patches, adding instance clustering to the framework to have the two largest clusters fed to the label generator and cleaner does help improve performance. Instance clustering essentially removes the fraction of uncertain patches from the data, which aids the proposed network in setting a decision boundary with high confidence. Similarly, the *label cleaner* module is crucial, which helps improve efficiency by around 6% by removing misclassified labels.

We also conducted experiments replacing transformer-based architectures with a simple Auto-encoder Label Generator (ALG) with seven fully connected layers [1024, 512, 256, 128, 256, 512, 1024] and an MLP-based Label Cleaner (MLC). This is to validate the transformer-based proposed architecture's use and effectiveness. The performance of this variant is 8.5% less than the proposed method with a transformer. Also, we observed that this variant had shown slightly better performance than the variant *w/o Label Generator*. This evidenced that the labeling based on projection and inverse project from latent space is adequate and effective, and the overall impact is enhanced using a transformer model. We set the maximum number of epochs to 200. However, we can reduce the number of epochs by using a proper convergence criterion (e.g., when the number of texture and non-texture labels stabilizes).

Table 4. Ablation studies on geometric feature selection for patch generation. *Cur* - curvature, *AZ* - azimuth angle, *SV* - surface variation, *EL* - elevation angle, *SI* - shape index, and *LD* - local depth

|     | [Cur, AZ, EL] | [LD, AZ, EL] | [SI, LD, Cur] | [SV, AZ, EL] | [SV, LD, AZ] | [SV, LD, Cur] | [SV, SI, AZ] | [SV, SI, Cur] | [SV, SI, LD] |
|-----|---------------|--------------|---------------|--------------|--------------|---------------|--------------|---------------|--------------|
| Pre | 56.8          | 54.3         | 67.3          | 54.2         | 50.1         | **68.1**      | 61.0         | 62.6          | 63.2         |
| Rec | 56.0          | 55.1         | **70.5**      | 52.7         | 52.6         | 69.6          | 60.7         | 63.0          | 64.5         |
| F1  | 57.1          | 54.7         | 68.5          | 53.4         | 51.3         | **70.0**      | 60.9         | 62.8          | 64.0         |

Table 5. Ablation study for proposed modules on *SHREC'18*. ALG stands for Auto-encoder-based Label Generator, *MLC* stands for MLP-based Label Cleaner and DL stands for discriminative learning.

|                       | Pre ↑ | Rec ↑ | F1 ↑ | mIoU ↑ |
|-----------------------|-------|-------|------|--------|
| *w/o Instance Clustering* | 66.2  | 64.8  | 66.0 | 70.1   |
| *w/o Label cleaner*   | 64.0  | 63.6  | 64.0 | 68.3   |
| *w/o Label Generator* | 58.4  | 60.5  | 61.5 | 64.0   |
| *w/o DL*              | 67.0  | 67.3  | 68.4 | 71.2   |
| *ALG + MLC*           | 61.7  | 60.1  | 61.5 | 65.6   |
| *Full network*        | **68.1** | **69.6** | **70.0** | **73.4** |

Table 6. Ablation study on grid size of feature patch generation on on *SHREC'18*. Bold represent the best performance and blue highlight represent the second best performance.

| Grid size | Pre ↑ | Rec ↑ | F1 ↓ | mIoU ↓ |
|-----------|-------|-------|------|--------|
| *8 x 8*   | 63.6  | 60.5  | 62.4 | 68.6   |
| *16 x 16* | 65.0  | 68.2  | 67.1 | 70.6   |
| *24 x 24* | **68.1** | **69.6** | 70.0 | 73.4 |
| *32 x 32* | 68.0  | 69.1  | **70.2** | **74.0** |
| *20 x 20* | 66.2  | 67.4  | 66.7 | 70.5   |

### 5.1. Parameters selection

We tested extensively the parameters that influence performance in the proposed approach.

**Grid size** is an important parameter since it determines the feature image size at each facet. Ideally, the grid size should cover a facet with sufficient surface area to determine whether the facet belongs to texture or not. Experimenting with various grid sizes, we found that 24 to 32 worked best with the proposed method, yielding superior results for both low- and high-resolution meshes. A small grid size does not adequately cover the surface area and performs poorly. Additionally, increasing the grid size has a border effect reducing the area of the segmented surface. Table 6 reports the performance of various grid sizes, revealing that grid sizes 24 and 32 provide superior performance compared to other grid sizes. As there is a slight performance difference between grid sizes 24 and 32, we chose grid size 24 for the experiment.

**Feature selection** is another important parameter that affects performance. We have tested multiple feature combinations to extract patches and checked the performance of the proposed approach. Since the texture pattern is a local variation on the surface, as expected, a combination of local depth, surface variations, and curvatures has shown better results. Table 4 summarizes the performance for different combinations. However, we observed that local depth plays an important role, and its combination with other geometric features has consistently shown better results. The other combinations related to surface variations, such as shape index, also show better results; however, they produce false positives in edge-like structures detected as texture.

## 6. Conclusions

In this paper, we proposed an original method for segmenting surfaces into textured and nontextured regions. Unlike previous techniques, which are limited to classification and retrieval and rely on human annotation for training networks, the proposed method is completely unsupervised. Extensive experiments on several datasets evidenced the segmentation capacity comparable to supervised methods. We plan to extend our work for a multi-class segmentation of textured surfaces.